\RequirePackage[T1]{fontenc}
\documentclass[letterpaper, 10 pt, conference]{ieeeconf}

\IEEEoverridecommandlockouts

\usepackage{algpseudocode}
\usepackage{algorithm}
\usepackage{booktabs}
\usepackage{graphicx}
\usepackage{cite}
\usepackage{hyperref}
\usepackage{amsmath}

\makeatletter
\let\PNPoriginalmakecaption\@makecaption
\long\def\@makecaption#1#2{%
  \begingroup
    \normalfont
    \let\footnotesize\normalsize
    \ifx\@captype\@IEEEtablestring
      \def\@captype{figure}%
      \PNPoriginalmakecaption{\fnum@table}{#2}%
    \else
      \PNPoriginalmakecaption{#1}{#2}%
    \fi
  \endgroup}
\renewcommand{\thetable}{\arabic{table}}
\def\fnum@table{Table~\thetable}
\makeatother

\title{\LARGE \bf
Uni-PNP: Unified Prehensile and Non-prehensile Policy Learning \\ for Generalizable Object Pose Reaching
}

\author{
\authorblockN{Wenxuan Zhu$^{1,2,*}$, Yuze Wang$^{1,2,*}$, Jiangran Lyu$^{1,2,*}$, Jiayi Chen$^{1,2,*}$, Yixin Zheng$^{2,3,4}$ \\
Mi Yan$^{1,2}$, Li Yi$^{2,5}$, Zhizheng Zhang$^{2}$, He Wang$^{1,2,\dagger}$}
\authorblockA{$^{1}$Peking University\quad $^{2}$Galbot\quad $^{3}$Institute of Automation, Chinese Academy of Sciences \\
$^{4}$Beijing Academy of Artificial Intelligence\quad $^{5}$Tsinghua University\\{\small $^{*}$Equal contribution}\quad \small $^{\dagger}$Corresponding authors}
}

\IEEEaftertitletext{%
  \hbox to \textwidth{\includegraphics[width=\textwidth]{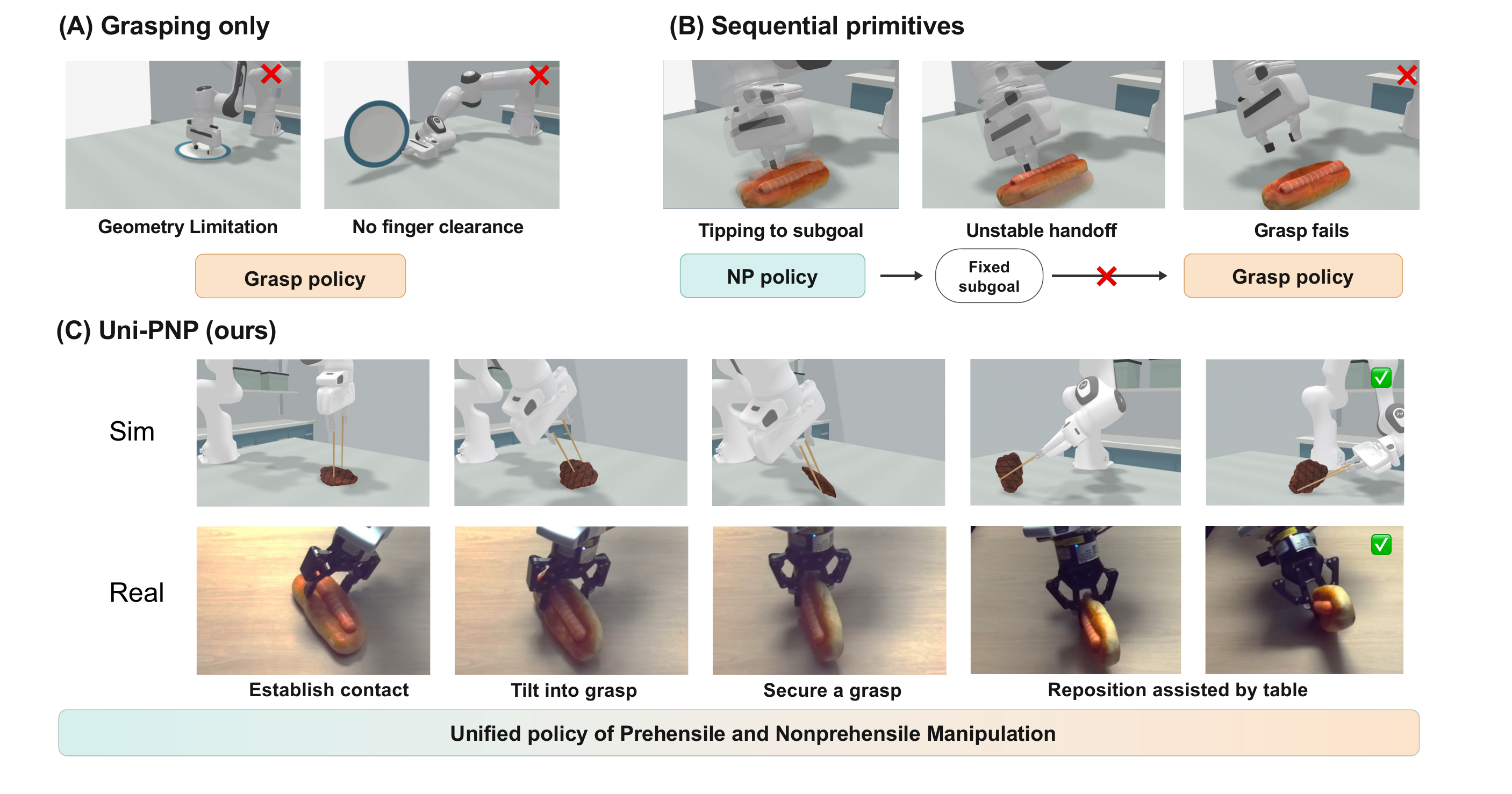}}%
  \vskip 0.04in
  \refstepcounter{figure}\label{fig:teaser}%
  \hbox to \textwidth{\parbox{\textwidth}{\normalsize\normalfont Fig.~\thefigure: Comparing different methods of prehensile and non-prehensile manipulation. (A) Grasping-only pipeline fails under geometric and contact constraints, and (B) sequential combination of non-prehensile and prehensile primitives fails when a fixed subgoal is unstable for the object. (C) Our policy (Uni-PNP) coordinates non-prehensile and prehensile interactions to tip objects into a grasp, instantly followed by a grasp secure. It also learns to leverage the table to reorient the object.
  }}%
  \vskip 8pt
}

\begin{document}

\maketitle
\thispagestyle{empty}
\pagestyle{empty}

\begin{abstract}
Object manipulation often requires combining non-prehensile actions with grasping. However, existing approaches typically develop these capabilities separately and integrate them by sequencing distinct primitives through prescribed intermediate objectives, limiting the coordination of contact motions and grasping. Their integration also remains sensitive to object and gripper geometry and depends on particular environmental supports, complicating real-world deployment. We present Uni-PNP, a unified policy that jointly learns prehensile and non-prehensile manipulation for object pose reaching across diverse objects and grippers. To achieve this, we first pretrain a geometry encoder to predict contact-induced object motion using a contact-dynamics dataset constructed from simulated interactions between diverse objects and grippers. We then train the policy in a two-stage reinforcement learning curriculum from graspable objects to ungraspable peers, encouraging the discovery of non-prehensile interactions and their integration with grasping. Experiments show 76.39\% success on hundreds of seen objects with grippers, and 57.91\% on unseen peers, outperforming the strongest pose-reaching baseline by 20.52 and 17.79 percentage points, respectively. We also analyze emergent behaviors, including nonprehensile and prehensile repositioning, grasp acquisition that requires grasping after tipping, and table-assisted reorientation while grasping, and transfer to the real-world scenario. Project page available at \href{https://lgswdn.github.io/Uni-PNP/}{lgswdn.github.io/Uni-PNP}.
\end{abstract}

\section{Introduction}

Robotic manipulation can be broadly divided into prehensile and nonprehensile modes. Prehensile manipulation moves an object while holding it in a stable grasp. Nonprehensile manipulation moves an object without grasping, through actions such as pushing or tipping. We often combine these skills in everyday activities. For instance, a phone lying flat on a table leaves little clearance for a direct grasp. We slide and tilt the phone while closing our fingers, creating room to secure and lift it. Nonprehensile contact creates access for grasping, while a secure grasp enables controlled lifting and repositioning. However, prior work typically develops prehensile and non-prehensile capabilities separately, and combines them either by preparing objects for a grasp using non-prehensile manipulation~\cite{preafford2024,gomp2026}, or by planning sequences of prehensile and non-prehensile primitives~\cite{hacmanpp2024,spin2025,adaptpnp2025}. Treating the two modes as separate skills limits their ability to learn unified manipulation behaviors.

Moreover, prior methods depend strongly on the geometry of the environment, object, and gripper. Several grasp-preparation strategies exploit environmental features such as edges and walls to obtain extrinsic dexterity~\cite{zhou2023extrinsic,preafford2024,kobashi2025occluded}, requiring suitable environmental supports. GOMP~\cite{gomp2026} uses a learned graspability field to guide non-prehensile manipulation toward configurations suitable for subsequent grasp execution. Furthermore, these methods are constrained to the gripper geometry seen in training, which is inconvenient for real-world deployment. Cross-gripper policies learn grasp poses from grasp synthesis ~\cite{adagr2021,graspgenx2026}. However, geometry influences manipulation beyond grasp pose selection: it also determines the contact interactions and resulting object motions. These geometry-dependent effects motivate jointly learning prehensile and non-prehensile manipulation across diverse objects and grippers.

We present Uni-PNP, a unified policy for prehensile and non-prehensile (PNP) object manipulation. Our policy reaches target object poses across diverse objects and grippers. To achieve this, we first construct a contact-dynamics dataset by sampling object--gripper contact configurations and recording the resulting object motions under gripper actions. We then pretrain a compact geometry encoder to predict these motions, learning a representation of geometry-dependent contact dynamics for the policy. Finally, we train the policy on these features through a two-stage reinforcement learning curriculum. The policy first learns grasping and repositioning with smaller, readily graspable objects. Training then continues on an object set whose initial configurations are challenging for direct grasping, encouraging the policy to discover non-prehensile interactions and combine them with grasping.

We evaluate Uni-PNP in simulation on 1,218 objects from this challenging set and 249 unseen objects from a different distribution, using grippers drawn from a training catalog of 200 geometries and the unseen Panda gripper. Uni-PNP achieves 76.39\% pose-reaching success on training objects and grippers, and 57.91\% on unseen objects with the unseen Panda gripper, outperforming the pose-reaching baselines across all settings. We complement these benchmarks with real-world experiments and an analysis of emergent manipulation skills, including tip-assisted grasping and table-assisted reorientation.

Our contributions are:
\begin{itemize}
    \item We introduce Uni-PNP, a unified policy that learns prehensile and non-prehensile manipulation jointly for object pose reaching.
    \item We construct a contact-dynamics dataset and propose motion-prediction pretraining to learn a compact geometry representation across varied objects and grippers.
    \item We demonstrate effective object pose reaching and generalization to unseen objects and grippers in simulation and real-world experiments.
\end{itemize}

\section{Related Work}

\subsection{Learning-based Non-Prehensile Manipulation}
Non-prehensile manipulation changes an object's pose through contacts such as pushing and pivoting ~\cite{mason1986,lynch1996,hou2018pivot}.  Recent learning-based methods unify non-prehensile manipulations by directly mapping geometric observations to actions~\cite{corn2024,unicorn2025,dywa2025,dapl2026}. CORN~\cite{corn2024} and HACMan~\cite{hacman2023} improve generalization across object shapes through contact-aware representations and end-to-end interaction learning, while other methods use parameterized primitives or policy decomposition~\cite{kim2023decomp}. Nevertheless, these approaches focus on non-prehensile repositioning itself; a non-prehensile policy alone does not secure the object for stable lifting. Grasp and post-grasp repositioning therefore remain separate objectives in most existing systems.

\subsection{Combining Prehensile and Non-Prehensile Manipulation}
\noindent\textbf{Grasp-Oriented Preparation and Affordance-Guided Manipulation.}
Many methods treat non-prehensile actions primarily as grasp preparation. Push--grasp systems coordinate pushing and grasping to improve grasp accessibility~\cite{dogar2010,xu2021pushgrasp,zeng2018}. Affordance-based methods predict regions or object configurations that support a subsequent grasp~\cite{preafford2024,gomp2026}, with language-conditioned affordances extending this idea to task-specific manipulation~\cite{yuan2025robopoint}. GOMP~\cite{gomp2026} learns a graspability field and directs non-prehensile motion toward configurations suitable for grasp execution. Several approaches obtain extrinsic dexterity from walls, edges, slots, or slopes~\cite{zhou2023extrinsic,preafford2024,kobashi2025occluded}, making execution dependent on suitable support geometry. Separating preparation from grasp execution introduces a handoff between the two stages, whose effectiveness depends on the intermediate object configuration and available support.

\noindent\textbf{Explicit Integration of Prehensile and Non-Prehensile Skills.}
Another line of work explicitly plans contact modes or sequences of manipulation skills. External pushes, shared grasping, and contact-mode planning coordinate finger contacts with environmental support~\cite{chavandafle2014,chavandafle2017,hou2020shared,cheng2021contact,cheng2022contact}. HACMan++~\cite{hacmanpp2024} selects spatially grounded motion primitives, AdaptPNP~\cite{adaptpnp2025} revises primitive sequences and intermediate object poses using execution feedback, and SPIN~\cite{spin2025} learns connectors between independently trained skills before distilling their planned trajectories into a policy. For skill-library approaches, the available primitives and transition conditions constrain the behaviors that can be composed. Contact-mode planners instead explore modeled contact interactions without necessarily requiring a predefined skill library. In contrast, Uni-PNP learns coordinated arm and gripper actions through a unified policy.

\subsection{Geometry Representations for Contact-Rich Control}
Point-cloud encoders such as PointNet~\cite{pointnet2017} provide geometric features for downstream learning. Self-supervised pretraining develops these features through correspondence matching~\cite{pointcontrast2020}, completion of occluded geometry~\cite{occo2021}, and masked prediction of discrete tokens, point geometry, or latent representations~\cite{pointbert2022,pointmae2022,point2vec2023}. For robotic control, PointPatchRL~\cite{pointpatchrl2024} combines masked reconstruction with reinforcement learning, while 4D visual pretraining learns features through future prediction~\cite{fourdp2025}. These objectives do not explicitly supervise the object motion induced by gripper contact.

\begin{figure*}[t]
\vspace*{5pt}
\centering
\includegraphics[width=\textwidth]{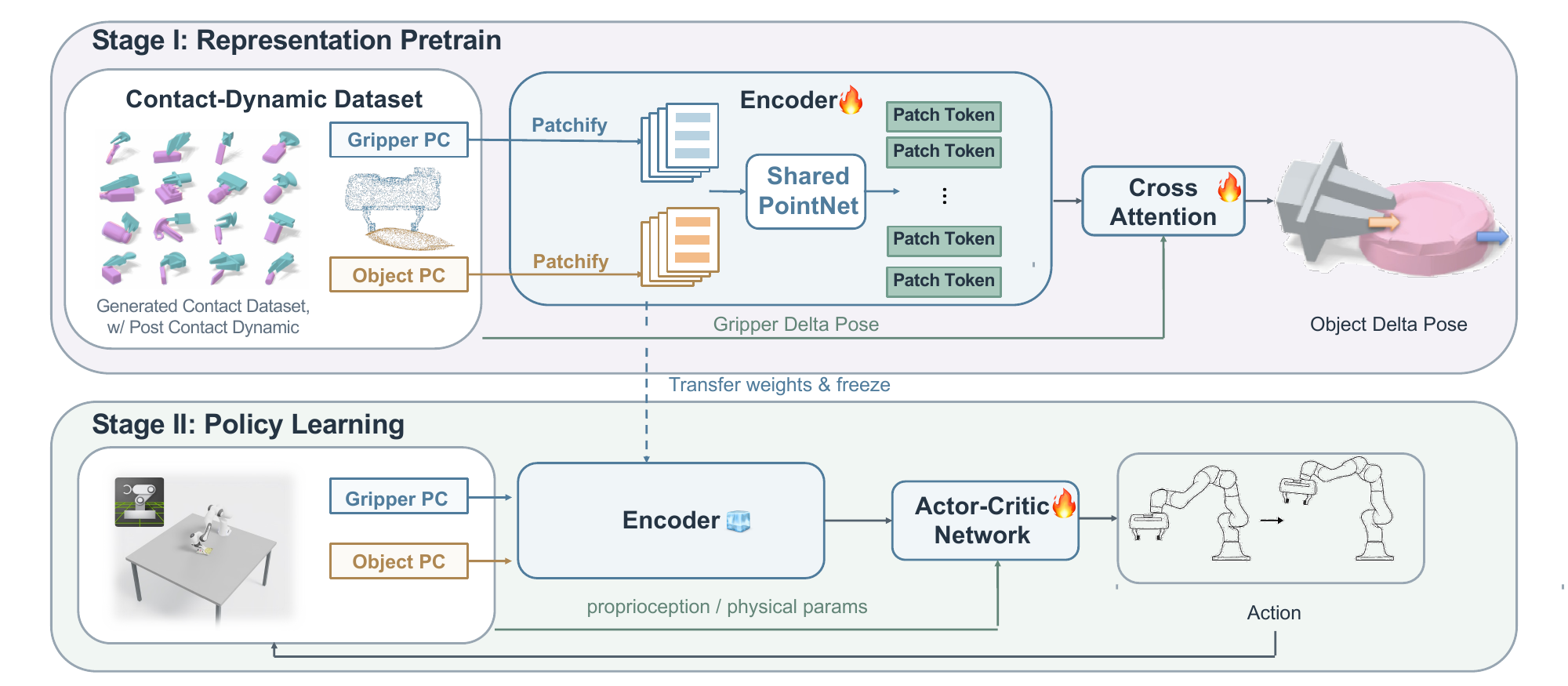}
\caption{Overview of Uni-PNP. We pretrain a geometry encoder to predict contact-induced object motion from object--gripper geometry and gripper actions. The pretrained encoder is then frozen during actor--critic policy learning with a two-stage curriculum.}
\label{fig:framework}
\end{figure*}

Contact-based representations focus on the geometry through which forces can be applied. CORN~\cite{corn2024} predicts contact between an object and the gripper, while UniCORN~\cite{unicorn2025} extends contact prediction to pairs of arbitrary geometries. Their supervision identifies where contact occurs but does not specify the motion it produces. DAPL~\cite{dapl2026} incorporates physical attributes and action-conditioned dynamics prediction to represent contact-induced scene motion. Its encoder is refined using trajectories collected by the RL policy, tying the learned representation to the task-specific interaction distribution explored during training. 

\section{Method}

Our goal is to achieve generalizable object pose reaching: bringing an object from an arbitrary stable initial pose to a specified target pose across different two-finger grippers. This requires combining non-prehensile contacts that reposition or reorient the object with prehensile actions that secure it for the final motion. To this end, we pretrain a geometry encoder to predict object motion induced by gripper contact under varying physical conditions. The learned representation conditions an actor--critic policy, which is optimized with a curriculum from ordinary grasping scenes to challenging object configurations. Fig.~\ref{fig:framework} summarizes our framework.

\begin{algorithm}
\caption{Contact-Dynamics Dataset Generation}\label{alg:cap}
\begin{algorithmic}
  \Require Centered shapes $\mathcal{O},\mathcal{G}$, dataset $\mathcal{C}$
  \State $s_O \sim p_s,\quad R_O \sim \mathcal{U}(\mathrm{SO}(3))$
  \State $T_O \leftarrow (R_O,t_O)$
  \State $p_O \sim \operatorname{SampleSurface}(T_O(s_O\mathcal O))$
  \State $p_G \sim \operatorname{SampleFingertip}(\mathcal G)$
  \State $R_G \sim \operatorname{SampleOrientation}(\mathcal G)$
  \State $t_G \leftarrow p_O-R_Gp_G,\quad T_G \leftarrow (R_G,t_G)$
  \If{$\operatorname{CollisionFree}(T_O,T_G,s_O\mathcal O,\mathcal G)$}
    \State $\phi \sim \operatorname{SamplePhysics}()$
    \State $T_O^{\mathrm{set}} \leftarrow \operatorname{Settle}(T_O,T_G,\phi)$
    \If{$\operatorname{InContact}(T_O^{\mathrm{set}},T_G)$}
      \State $\Delta T_G \sim \operatorname{SampleMotion}()$
      \State $\Delta T_O \leftarrow \operatorname{Simulate}(T_O^{\mathrm{set}},T_G,\phi,\Delta T_G)$
      \State $\mathcal{C} \leftarrow \mathcal{C} \cup \{(T_O^{\mathrm{set}},T_G,\phi,\Delta T_G,\Delta T_O)\}$
    \EndIf
  \EndIf
\end{algorithmic}
\end{algorithm}

\subsection{Representation Pretraining}

We pretrain our geometry encoder on simulated contact interactions before optimizing the manipulation policy. Each transition pairs the observed object--gripper geometry and physical conditions with the object motion induced by the gripper, exposing the representation to the consequences of contact.

\noindent\textbf{Contact-Dynamics Dataset Generation.}
Algorithm~\ref{alg:cap} summarizes the data-generation procedure. We use objects from DexGraspNet~\cite{dexgraspnet} and generate grippers with diverse geometries following ZETA~\cite{zeta2026}. For each object--gripper pair, we sample an object scale $s_O$ and rotation $R_O$ and place the object on the ground plane. We then sample an object surface point $p_O$, a gripper fingertip point $p_G$, and a gripper orientation $R_G$, and translate the gripper so that the selected points coincide. Configurations that penetrate the object or ground are rejected.

For each accepted configuration, we sample the physical parameters
  \[
  \phi =
  \left(
  m_O,\; m_G,\; \mu_O,\; \mu_G,\; \mu_{\mathrm{ground}}
  \right),
  \]
where $m_O$ and $m_G$ denote the object and gripper masses, and $\mu_O$, $\mu_G$, and $\mu_{\mathrm{ground}}$ denote the corresponding friction coefficients. We first let the initialized scene settle for $t_{\mathrm{settle}}$ physics steps and retain configurations that remain in contact. We then sample a gripper pose change $\Delta T_G$, execute it uniformly over $t_{\mathrm{motion}}$ physics steps, and record the resulting object pose change $\Delta T_O$ from the settled state.

Each valid transition defines the prediction task
  \[
  f\!\left(
  X_O,\; X_G,\; \phi,\; \Delta T_G
  \right)
  \;\longrightarrow\;
  \Delta T_O,
  \]
where $X_O$ and $X_G$ are the object and gripper point clouds expressed in the object-centered coordinate frame.
Here, $p_s$ is the object-scale distribution, and $t_O$ places the lowest object point on the ground with zero horizontal translation. 

\begin{figure}[t]
\vspace*{5pt}
\centering
\includegraphics[width=\columnwidth]{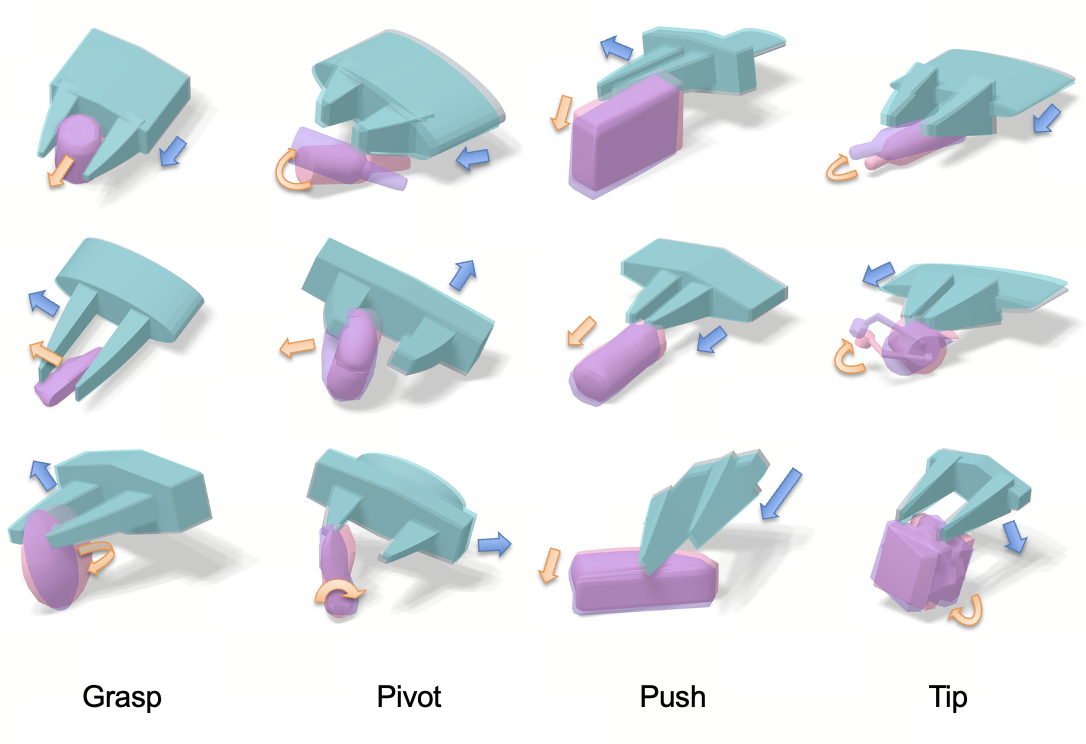}
\caption{Selected contact interactions represented in the pretraining dataset with various generated grippers.}
\label{fig:contact_dataset_skills}
\end{figure}

\noindent\textbf{Encoder Architecture.}
Following CORN~\cite{corn2024}, we select $M$ patch centers from each point cloud using farthest-point sampling (FPS) and group neighboring points into local patches using K-nearest neighbors (KNN). This yields $2M$ patches in total. For each point in a patch, our encoder constructs a seven-dimensional feature
  \[                          
  (x,y,z,p_x,p_y,p_z,c),
  \] 
where \((x,y,z)\) are the point coordinates relative to the patch center, \((p_x,p_y,p_z)\) are the patch-center coordinates in the object-centered frame, and \(c\) is a binary identifier distinguishing object and gripper points. A shared pointwise MLP, max pooling, and a patch-level MLP embed each patch into a $d$-dimensional token.

\noindent\textbf{Dynamics Pretraining Objective.}
The dynamics pretraining module conditions the $2M$ geometry tokens on the relative gripper position, gripper motion $\Delta T_G$, and physical parameters $\phi$ through cross-attention. The updated object and gripper features are pooled and passed to an MLP to predict the object pose change $\widehat{\Delta T}_O=(\widehat{\Delta t}_O,\widehat{\Delta r}_O)$.

We minimize a pose-regression loss averaged over each minibatch. The translation term is
  \[
  \mathcal{L}_{\mathrm{trans}}=
  \left\|\widehat{\Delta t}_O-\Delta t_O\right\|_2,
  \]
and the rotational term is
  \[
  \mathcal{L}_{\mathrm{rot}}=1-
  \left\langle q(\widehat{\Delta r}_O),q(\Delta r_O)\right\rangle^2,
  \]
where $q(\cdot)$ denotes the unit-quaternion representation. The total
objective is
  \[
  \mathcal{L}_{\mathrm{dyn}}=\lambda_t\mathcal{L}_{\mathrm{trans}}
  +\lambda_r\mathcal{L}_{\mathrm{rot}}.
  \]
The encoder is optimized jointly with the dynamics prediction head, and the resulting
representation is subsequently used by the policy.

\subsection{PNP Policy Training via Reinforcement Learning}

Given the pretrained geometry representation, we train the downstream policy with reinforcement learning in simulation and a curriculum that progresses from basic prehensile manipulation to challenging initial configurations. We optimize the actor--critic policy with Proximal Policy Optimization (PPO)~\cite{schulman2017ppo} using 8,192 parallel environments.

\noindent\textbf{Ungraspable Objects.}
We select ungraspable objects by sampling an object scale and initial pose, then using GraspGen~\cite{graspgen2025} to generate grasp candidates with confidence greater than $\tau_{grasp}$. Each candidate is checked for collisions between the gripper and the object or the ground plane. We retain objects with at least one generated candidate but no collision-free candidate, selecting initial configurations that are challenging for direct grasping.  We further use StereoVLA~\cite{stereovla2026} to filter out objects that are graspable in simulation with Panda gripper, guarding against cases where GraspGen favors invalid grasp candidates for very flat objects. The sampled object scale is fixed throughout policy training and evaluation.

\noindent\textbf{Policy Architecture.}
At each timestep, our policy observes the object and gripper point clouds, a goal pose, physical parameters, and the robot proprioceptive state. The actor outputs an eight-dimensional relative action consisting of seven joint-position increments and one gripper-opening increment.

Our policy first translates both point clouds by $-t_O$ to express them in an object-centered frame, then feeds them into the geometry encoder to produce $2M$ tokens. Following CORN~\cite{corn2024}, the remaining observations, together with $t_O$, are projected into query features that attend to these tokens through cross-attention. The fused representation is passed through MLP heads to produce the actor output and critic value.

\noindent\textbf{Reward Design.}
We follow the reward design of DAPL~\cite{dapl2026}, which consists of a sparse task success term, a contact term that rewards contact between the gripper and the object, and a pose-reaching term that rewards goal-directed object motion.

\noindent\textbf{Curriculum Policy Training.}
We train the policy with a two-stage curriculum. In the first stage, the policy is optimized for $5000$ iterations on the DGN object distribution with smaller object scales. The resulting checkpoint initializes the second stage, where training continues for $10^4$ iterations on the selected ungraspable objects. The first stage develops basic grasping and pose-reaching skills, while the second stage encourages the policy to combine non-prehensile interactions with grasping.

\section{Experiments}

We evaluate the effectiveness of contact-dynamics pretraining and its generalization across objects and grippers.

\subsection{Simulation Experiments}

\noindent\textbf{Setup and Baselines.}
We evaluate on the selected ungraspable objects and a set of unseen and ungraspable objects selected from Objaverse~\cite{objaverse2023}. Each object is tested for ten episodes. An episode is considered successful if the object maintains within 0.05 m and 0.1 rad of the target pose for 1 second. In addition to the training grippers, we use the Panda gripper to evaluate zero-shot gripper transfer. We compare our method with the following two types of baselines:

\begin{itemize}
    \item \textbf{Manipulation policies:} GOMP is a released policy for making objects graspable through learned non-prehensile manipulation. We evaluate its reference-free inference procedure using its own graspability and grasp-pose predictions.
    \item \textbf{Geometric representations:} We compare against Point2Vec and UniCORN. UniCORN (original) uses the released encoder~\cite{unicorn2025}, while UniCORN (retrained) is pretrained on our object set and generated grippers with the original contact objective. To ablate pretraining, we include PointNet with contact pretraining following UniCORN and PointNet (E2E), which trains the encoder jointly with the policy from scratch.
    
\end{itemize}

For the representation-based methods except PointNet (E2E), we freeze the pretrained encoder and train the same downstream actor--critic architecture on its features. We report object pose-reaching success for these methods. We evaluate GOMP on the same object sets with the Panda gripper, its only supported gripper, and report success in grasping, lifting, and holding. Object and gripper transfer are evaluated zero-shot.

\noindent\textbf{Main Results.}
Table~\ref{tab:simulation_success} summarizes performance across object and gripper settings. The training-gripper columns report performance with grippers drawn from the training catalog; the Panda columns report zero-shot transfer to the Panda gripper.

Uni-PNP outperforms the evaluated pose-reaching baselines, achieving 76.39\% success on ungraspable objects with training grippers, 20.52 percentage points above UniCORN (retrained). This advantage persists under gripper transfer, with 66.57\% success on Panda. Uni-PNP achieves these results with a compact PointNet encoder pretrained on contact dynamics.

\begin{table}[t]
\vspace*{5pt}
\centering
\renewcommand{\arraystretch}{1}
\setlength{\tabcolsep}{3pt}
\begin{tabular*}{\columnwidth}{@{\extracolsep{\fill}}lcccc@{}}
\toprule
& \multicolumn{2}{c}{\shortstack{Ungraspable\\objects}} & \multicolumn{2}{c}{\shortstack{Unseen\\Objaverse objects}} \\
\cmidrule(lr){2-3}\cmidrule(lr){4-5}
Method & \shortstack{Training\\grippers} & Panda & \shortstack{Training\\grippers} & Panda \\
\midrule
GOMP* & -- & 12.33 & -- & 11.04 \\

PointNet (E2E) & 37.67 & 19.43 & 37.55 & 16.91 \\
UniCORN (original) & 2.64 & 2.25 & 4.58 & 4.78 \\
UniCORN (retrained) & 55.87 & 42.86 & 50.08 & 40.12 \\
Point2Vec & 50.64 & 34.63 & 46.99 & 32.05 \\
\textbf{Ours} & \textbf{76.39} & \textbf{66.57} & \textbf{69.48} & \textbf{57.91} \\
\bottomrule
\end{tabular*}
\caption[Baseline comparison.]{\textbf{Baseline comparison.} Uni-PNP outperforms the pose-reaching baselines across object and gripper settings. GOMP* reports success in grasping, lifting, and holding; all other methods report object pose-reaching success.}
\label{tab:simulation_success}
\end{table}

\begin{figure}[t]
\vspace*{5pt}
\centering
\includegraphics[width=\columnwidth]{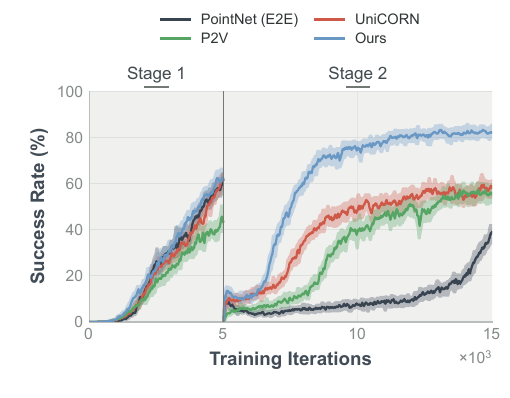}
\caption{Training success across the two-stage curriculum. The vertical line marks the stage transition at 5,000 iterations. UniCORN and P2V denote UniCORN (retrained) and Point2Vec, respectively.}
\label{fig:training_curve}
\end{figure}

\noindent\textbf{Training Efficiency.}
Fig.~\ref{fig:training_curve} compares learning progress under the same two-stage curriculum. Our policy improves more quickly after the stage transition and reaches higher training success, indicating that contact-dynamics pretraining supports efficient learning on the challenging object set.

\subsection{Ablation Study}

\noindent\textbf{Ablation of Architecture and Pretraining.} We ablate the encoder architecture and pretraining strategy in Table~\ref{tab:architecture_pretraining}. Dynamics pretraining increases PointNet's success rate from 10.89\% to 76.39\%, while UniCORN achieves similar success rates with contact and dynamics pretraining. With dynamics pretraining, the compact PointNet encoder outperforms UniCORN by 21.28 percentage points.

\noindent\textbf{Ablation of Curriculum Learning.} We compare the adopted object-difficulty curriculum with two alternatives. Training directly on the full set of ungraspable objects without a curriculum remains near zero success throughout 15,000 iterations, indicating that the policy rarely discovers the coordinated contact sequences from scratch. We also train first on stable target poses and then switch to arbitrary targets. The policy reaches 69.98\% on the stable-pose stage, but fails to make progress after the switch. Stable targets admit a shortcut in which the policy closes the gripper and relies on non-prehensile table contact, without acquiring the grasped pose-control behavior needed for arbitrary goals. In contrast, our curriculum first uses simpler objects to establish prehensile grasping and repositioning, then introduces configurations that require non-prehensile repositioning. The comparison shows that effective curriculum design must order the interaction difficulties, rather than only simplify the target-pose distribution.

\begin{table}[t]
\vspace*{5pt}
\centering
\renewcommand{\arraystretch}{1.1}
\setlength{\tabcolsep}{6pt}
\begin{tabular*}{\columnwidth}{@{\extracolsep{\fill}}lcc@{}}
\toprule
& \multicolumn{2}{c}{Pretraining} \\
\cmidrule(lr){2-3}
Architecture & Contact & Dynamics \\
\midrule
PointNet & 10.89 & \textbf{76.39} \\
UniCORN & 55.87 & 55.11 \\
\bottomrule
\end{tabular*}
\caption[Ablation of architecture and pretraining.]{\textbf{Ablation of architecture and pretraining.} PointNet with dynamics pretraining is our encoder; UniCORN with contact pretraining corresponds to UniCORN (retrained).}
\label{tab:architecture_pretraining}
\end{table}

\subsection{Skill Emergence of Unified Prehensile and Non-Prehensile Manipulation}

\begin{figure}[t]
\vspace*{5pt}
\centering
\includegraphics[width=\columnwidth]{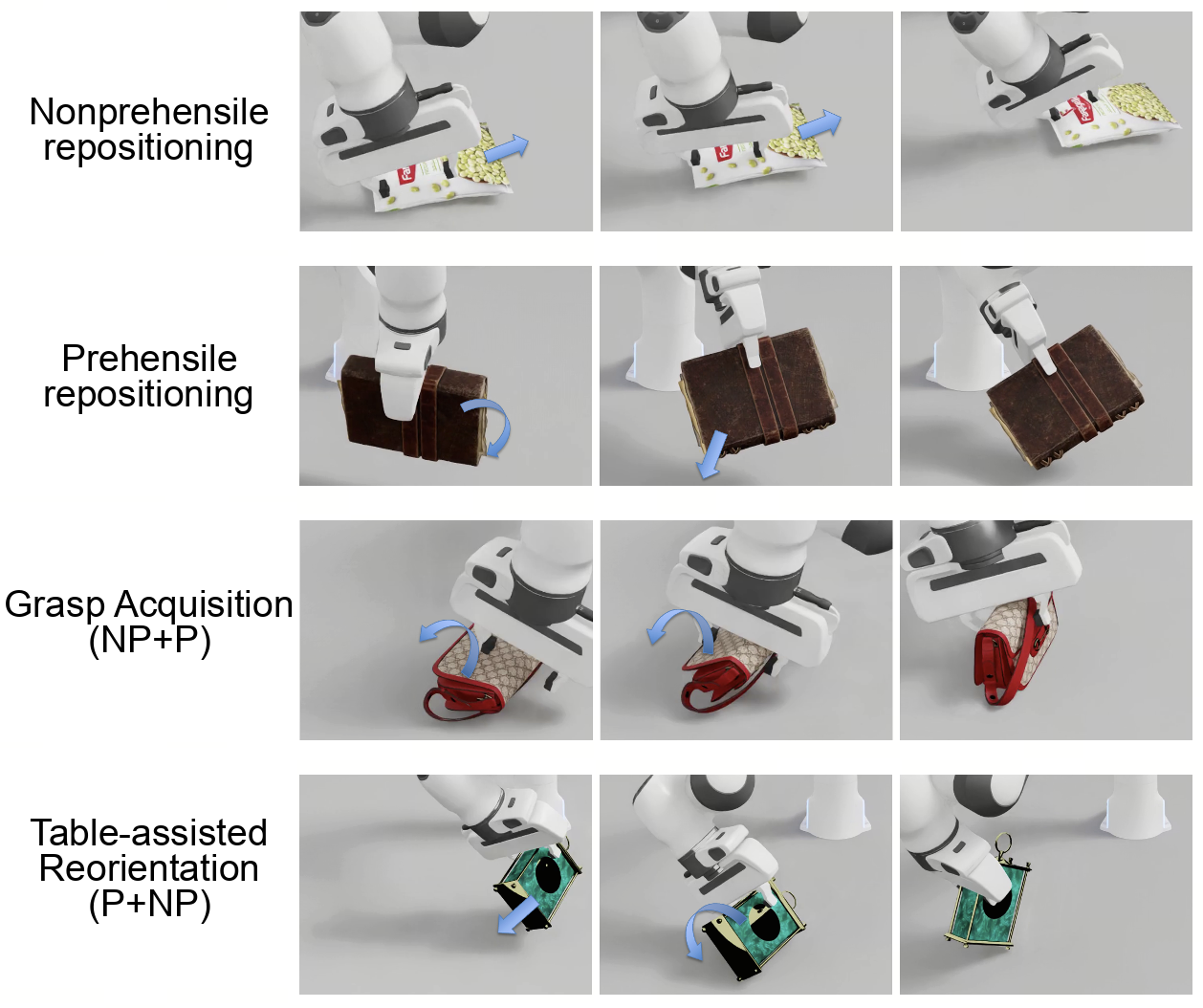}
\caption{Emergent prehensile and non-prehensile manipulation skills. Rows show NP only, P only, NP+P, and P+NP, respectively.}
\label{fig:emergent_skills}
\end{figure}

\begin{figure}[t]
\vspace*{5pt}
\centering
\includegraphics[width=\columnwidth]{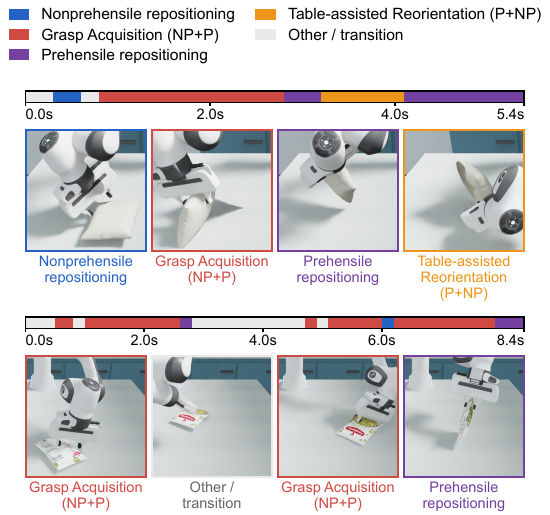}
\caption{Skill composition in two representative qualitative rollouts. Colored bars indicate manipulation phases.}
\label{fig:skill_timelines}
\end{figure}

\begin{figure*}[t]
\vspace*{5pt}
\centering
\includegraphics[width=\textwidth]{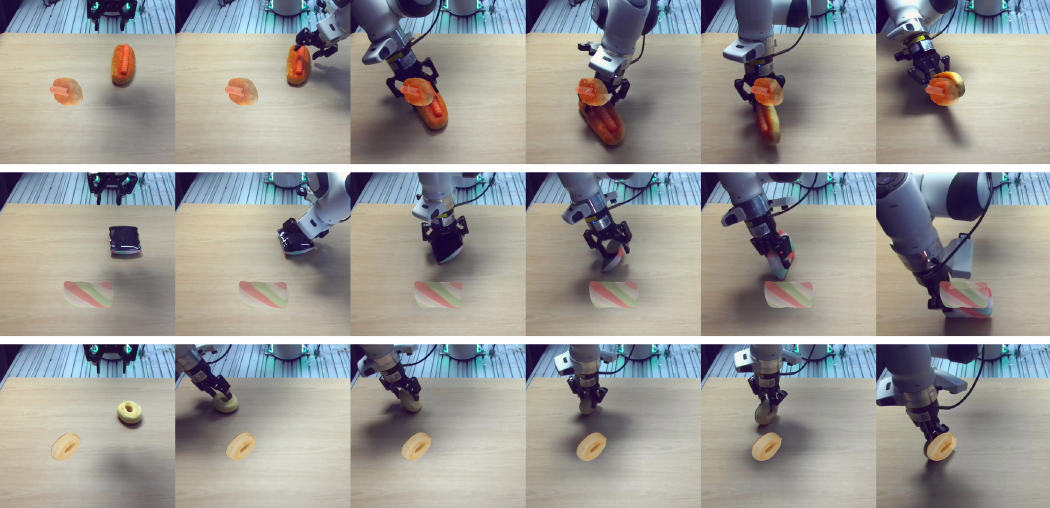}
\caption{Qualitative real-world object pose-reaching sequence. The goal pose is shown transparently.}
\label{fig:real_world_sequence}
\end{figure*}

To examine how the learned policy combines prehensile and non-prehensile manipulation, we study four types of behavior. \emph{Non-prehensile repositioning} uses pushing and pivoting to change an object's position and orientation on the table. \emph{Prehensile repositioning} holds the object and moves it toward the goal through arm and wrist motion. \emph{Grasp Acquisition (NP+P)} is a skill where grasping follows tightly after non-prehensile tipping: the policy tips an object against the table to expose an accessible region, then closes the gripper to secure it. For some objects, the other finger provides support during tipping and then slides downward to complete the grasp. \emph{Table-assisted Reorientation (P+NP)} changes the object's orientation within an established grasp. In particular, the policy can maintain its grip while pressing the object against the table, using table contact to obtain a new orientation relative to the fingers. Fig.~\ref{fig:emergent_skills} illustrates these behaviors.

Beyond switching between skills, the policy can also revise an acquired grasp. A grasp that supports lifting may constrain the rotations needed to reach the goal; releasing the object onto the table and regrasping it provides a new configuration for reorientation.

Fig.~\ref{fig:skill_timelines} shows how the policy combines these skills in two qualitative rollouts. The first rollout includes table-assisted reorientation; in the second, the object flips onto its opposite face before the policy regrasps it.

To quantify these behaviors, we analyze successful simulated episodes on ungraspable objects with the training grippers. We segment trajectories using gripper--object and object--table contacts, gripper closure, and object motion in the world and gripper frames. Table-supported sliding and turning identify non-prehensile repositioning, while closure, fingertip contact, and preparatory tipping identify grasp acquisition. A stable grasp followed by maintained contact identifies prehensile repositioning. Table-assisted reorientation is identified by a sustained relative rotation within an established stable grasp that settles into a new object--gripper relative pose. 

\begin{table}[t]
\vspace*{5pt}
\centering
\small
\setlength{\tabcolsep}{3pt}
\begin{tabular*}{\columnwidth}{@{\extracolsep{\fill}}lrr@{}}
\toprule
Skill & \shortstack{Occurrence\\(\%)} & \shortstack{Time share\\(\%)} \\
\midrule
\shortstack[l]{non-prehensile repositioning} & 76.71 & 9.00 \\
\shortstack[l]{prehensile repositioning} & 99.18 & 21.82 \\
\shortstack[l]{grasp acquisition} & 100.00 & 40.70 \\
\shortstack[l]{table-assisted reorientation} & 18.90 & 1.53 \\
\bottomrule
\end{tabular*}
\caption[Manipulation skills in successful episodes.]{\textbf{Manipulation skills in successful episodes.} Free motion and transitions account for the remaining time.}
\label{tab:skill_usage}
\end{table}

Table~\ref{tab:skill_usage} reports the percentage of successful episodes containing each skill and the share of episode time spent on it. Most manipulation time is spent acquiring and controlling grasps, while non-prehensile repositioning appears across many successful episodes. Table-assisted reorientation serves as a brief adjustment within these longer sequences.

\subsection{Real-World Experiments}

Our real-world setup follows DAPL~\cite{dapl2026}, with a Franka Research 3 robot connected to Robotiq 2F85 gripper, and three Intel RealSense cameras. XMem~\cite{xmem2022} tracks the object segmentation, and FoundationPose~\cite{foundationpose2024} estimates its 6D pose. We transfer the simulation-trained policy through teacher--student distillation. To support more refined control, we execute policy commands through the low-level controllers provided by CRISP~\cite{crisp2025}. We limit the end-effector's linear and angular speeds to $0.1\,\mathrm{m/s}$ and $1\,\mathrm{rad/s}$, respectively.

We examine real-world behavior through qualitative object pose-reaching trials. Fig.~\ref{fig:real_world_sequence} shows a representative sequence with object repositioning, grasp acquisition, and object reorientation. We also attempted these tasks using teleoperation, but did not achieve similar behaviors in our trials. These tasks require fine-grained coordination between gripper and arm actions, motivating our approach to learning coordinated manipulation in simulation.

\section{CONCLUSIONS AND LIMITATIONS}

In this work, we proposed \textbf{Uni-PNP}, a unified policy for general object pose reaching that combines prehensile and non-prehensile manipulation. We introduced a contact-dynamics dataset and motion-prediction pretraining to learn a compact geometry representation, and used a two-stage reinforcement learning curriculum to transfer basic grasping and pose control to challenging ungraspable objects. Our approach achieves effective pose reaching across diverse objects and grippers, generalizes to unseen objects and grippers, and exhibits coordinated behaviors including repositioning, grasp acquisition, and table-assisted reorientation in simulation and real-world trials.

Our approach has several limitations. Firstly, the contact-dynamics dataset and policy training currently focus on tabletop interactions without introducing diverse environments or cluttered scenes. Secondly, our generated grippers are limited to two-finger designs. Generalization to hands with multiple fingers remains an open question. Finally, we focus on basic object pose reaching and do not consider tasks with semantic information. Future work could integrate the policy with planning-based methods for long-horizon tasks.

\bibliographystyle{IEEEtran}
\bibliography{references}

\end{document}